\documentclass[sigconf, 9pt]{acmart}
\AtBeginDocument{%
  }

 \copyrightyear{2025} 
\acmYear{2025} 
\setcopyright{cc}
\setcopyright{cc}
 \setcctype{by}
 \acmConference[BUILDSYS '25]{The 12th ACM International Conference on
 Systems for Energy-Efficient Buildings, Cities, and Transportation}{November
 19--21, 2025}{Golden, CO, USA}
 \acmBooktitle{The 12th ACM International Conference on Systems for
 Energy-Efficient Buildings, Cities, and Transportation (BUILDSYS '25), November
 19--21, 2025, Golden, CO, USA}
 \acmDOI{10.1145/3736425.3771958}
 \acmISBN{979-8-4007-1945-5/2025/11}

\usepackage[export]{adjustbox} 
\usepackage{comment}

\usepackage{amssymb}
\usepackage{amsfonts}
\usepackage{gensymb}
\usepackage{subcaption}
\usepackage[table]{xcolor}
\usepackage{colortbl}
\usepackage{float}      
\usepackage{chngcntr}   
\usepackage{multirow,arydshln}
\usepackage{pifont}    
\usepackage{tabularx} 
\counterwithin{figure}{section}
\counterwithin{table}{section}
\usepackage{booktabs}      
\usepackage{adjustbox}     
\usepackage{multirow}      
\usepackage{makecell} 

\usepackage{array}      
\newcolumntype{L}[1]{>{\raggedright\arraybackslash}p{#1}}  
\begin{document}



\title{\vspace{-0.6em}{Evaluating Spatio-Temporal Forecasting Trade-offs Between Graph Neural Networks and Foundation Models}}

\author{
\fontsize{9pt}{9pt}\selectfont Ragini Gupta$^{*}$, Naman Raina$^{*}$, Bo Chen$^{*}$, Li Chen$^{'}$, Claudiu Danilov$^{+}$, Josh Eckhardt$^{+}$, Keyshla Bernard$^{+}$, Klara Nahrstedt$^{*}$ \\
$^{*}$University of Illinois Urbana-Champaign, USA, $^{'}$University of Louisiana at Lafayette, USA,
$^{+}$Boeing Research and Technology, USA\\
\{raginig2, namanr2, boc2, klara\}@illinois.edu, \{li.chen\}@louisiana.edu, \{cdanilov, josh.d.eckhardt, keyshla.j.bernard\}@boeing.com
}
\renewcommand{\shortauthors}{}
\begin{abstract}
\vspace{-0.2cm}
Modern IoT deployments for environmental sensing produce high volume spatiotemporal data to support downstream tasks such as forecasting, typically powered by machine learning models. While existing filtering and strategic deployment techniques optimize collected data volume at the edge, they overlook how variations in sampling frequencies and spatial coverage affect downstream model performance. In many forecasting models, incorporating data from additional sensors denoise predictions by providing broader spatial contexts. This interplay between sampling frequency, spatial coverage and different forecasting model architectures remain underexplored. This work presents a systematic study of forecasting models - classical models (VAR), neural networks (GRU, Transformer), spatio-temporal graph neural networks (STGNNs), and time series foundation models (TSFMs: Chronos Moirai, TimesFM) under varying spatial sensor nodes density and sampling intervals using real-world temperature data in a wireless sensor network. Our results show that STGNNs are effective when sensor deployments are sparse and sampling rate is moderate, leveraging spatial correlations via encoded graph structure to compensate for limited coverage. In contrast, TSFMs perform competitively at high frequencies but degrade when spatial coverage from neighboring sensors is reduced. Crucially, the multivariate TSFM Moirai outperforms all models by natively learning cross-sensor dependencies. These findings offer actionable insights for building efficient forecasting pipelines in spatio-temporal systems. All code for model configurations, training, dataset, and logs are open-sourced for reproducibility.\footnote{\url{https://github.com/UIUC-MONET-Projects/Benchmarking-Spatiotemporal-Forecast-Models}} 
\vspace{-0.25cm}
\end{abstract}

\vspace{-0.2cm}
\ccsdesc[500]{Information systems~Spatial-temporal systems}
\ccsdesc[500]{Computing methodologies~Machine learning}
\keywords{Spatio-temporal forecasting, Graph Neural Networks, Time Series Foundation Models}

\maketitle
\vspace{-0.5cm}
\section{Introduction}
\vspace{-0.1cm}
Urban IoT systems such as environmental monitoring platforms rely on short-term weather forecasting (2–6h ahead) techniques to support decision making in mobility planning and environmental control. Densely deployed meteorological sensors such as temperature, pressure or humidity in a wireless sensor network offer rich spatial and temporal coverage but raises questions on how sampling frequency and inter-sensor (node) correlations affect downstream forecasting, an aspect often overlooked by edge-side data acquisition methods like adaptive sampling and filtering \cite{adaptive}. Forecasting approaches range from statistical  models (Autoregressive, Recurrent Neural Network-RNN) to modern architectures (Transformers, Spatio-Temporal Graph Neural Networks- STGNN), each with different sensitivities: Transformers excel at high-frequency data, while RNNs may overfit or underfit depending on resolution \cite{kdd}. STGNNs explicitly model graph structures to capture spatiotemporal dependencies \cite{stgnn, stgnn2}. 
Most recently, Time Series Foundation Models (TSFMs) like Chronos \cite{ansari2024chronos}, Moirai \cite{moirai}, TimesFM \cite{timesFM} have emerged, leveraging large-scale pretraining to enable strong zero-shot forecasting across diverse domains. 
Unlike language, time-series data is highly heterogeneous, making forecasting complex in domains like temperature where dynamic, confounding variables are inherent. TSFMs eliminate ad-hoc modeling by providing a single pre-trained model that works across tasks with zero-shot inference or minimal fine-tuning, demonstrating strong single-sensor univariate forecasting \cite{buildsys24}, but their effectiveness in multi-sensor spatiotemporal settings remains an open question. This leads to a critical question: \textbf {How do explicit graph-based structures (STGNNs) compare against pretrained Time Series Foundation Models, for short-term temperature forecasting under different spatio-temporal conditions?} To answer this, we compare STGNNs with VAR, RNNs, Transformers, and two state of the art TSFM models —Moirai \cite{moirai}, Chronos \cite{ansari2024chronos}, TimesFM \cite{timesFM}—using a 25-node IoT meterological dataset from New Mexico \cite{iobt, iobt2}. We vary \textbf{(1)}: Temporal resolution through downsampling (5-60 minute intervals) to assess model sensitivity to sampling rates, and \textbf{(2)}: Number of nodes (8, 16, 25) to examine robustness across spatial densities. This setup enables comparison of graph vs. non-graph models under varied spatiotemporal conditions. Our findings show that STGNNs outperform classical and Transformer based models when spatial correlations are strong or data is sparse, while TSFMs perform competitively in uniform, high-frequency settings with low spatial variability. This highlights the need to align model choice with data topology and sensing strategy. 
\vspace{-0.45cm}
\section{Related Works}
\vspace{-0.2cm}
\noindent{\textbf{\textit{Graph Neural Networks for time-series forecasting.}} Time-series forecasting progressed from statistical ARIMA to deep learning, with early sequence models such as Long Short-Term Memory (LSTM) \cite{lstm3} and Gated Recurrent Units (GRU) (an RNN variant) \cite{gru2} captured temporal patterns but largely ignored spatial relationships. Spatio-Temporal GNNs (STGNNs) \cite{weather3} model spatial–temporal dependencies by treating sensors as graph nodes, with graphs built from geographic proximity or cross-sensor similarity (modalities/meteorology). Refinements include STGCN \cite{stgcn} and Graph WaveNet \cite{wavenet}, and later attention/Transformer-based STGNNs \cite{attention4}. Despite these advances, such models rely heavily on training data and often generalize poorly to unseen settings.

\noindent{\textbf{\textit{Time-Series Foundation Models for forecasting.}} Inspired by the transformative success of Large Language Models (LLMs), Time-Series Foundation Models (TSFMs) have emerged as a promising path toward generalizable time-series analysis, demonstrating strong few-shot and zero-shot generalization across domains like cyber-physical systems \cite{tsfm4}, building energy analytics \cite{tsfm2, buildsys24}, and human activity recognition \cite{tsfm3}.  These models, based on adapted transformer architectures, are pre-trained on large, diverse corpora of time-series data empowering them to perform core tasks like forecasting and anomaly detection.  Contemporary TSFMs like MOMENT \cite{moment} and TimesFM \cite{timesFM} aim to generalize across datasets and  modalities, but remain limited to a fixed set of tasks. Although these models support covariate inputs as exogenous variables, their integration methods such as flattening multivariate sequences through variate embeddings or using similarity-based attention, have failed to show consistent performance gains as the transformer encoder treats these covriates as extra input tokens alongside the past target history to form context representation, thus, failing to capture the inter-variable relationship.   This limitation becomes critical in spatio-temporal environments such as urban sensor networks, where data heterogeneity across sensor types, locations, and sampling resolutions creates complex inter-dependencies that their design is ill-equipped to handle. In contrast, newer generation of true multivariate TSFMs, such as Moirai \cite{moirai}, are built on the core assumption of the transformer that all channels in a multivariate series are inherently correlated and interdependent, however, their performanc not studied. Prior work has examined TSFMs, STGNNs \cite{stgnn, stgnn2}, and sequential models \cite{kdd} in isolation, leaving a critical need for a controlled, cross-paradigm comparison. We rigorously quantify the performance of TSFMs against spatio-temporal Graph Neural Networks, revealing their trade-offs under varied sampling rates and spatial node coverage (or node densities). This analysis establishes a critical empirical baseline, providing essential insight into the practical challenges of data heterogeneity and precisely defining the necessity for effective multi-variate integration in future TSFMs.
\vspace{-0.8cm}
\section{Overview of Time Series Forecasting Models}
\vspace{-0.2cm}
We group models into 3 categories: Classical, TSFMs, and STGNNs. \noindent\textbf{\textit{A) Classical baseline models:}} We include baseline models: Vector AutoRegression (VAR)—a multivariate AR model, Gated Recurrent Unit (GRU)—a sequential RNN variant, and Transformer~\cite{transformer}—which models temporal dependencies using self-attention. 

\noindent\textbf{\textit{B) 
 Time Series Foundation Models (TSFMs):}} TSFMs are large pretrained models designed for zero-shot or transfer forecasting. We evaluate three TSFM models: Moirai,  TimesFM, and Chronos. Table~\ref{tab:tsfm-base-attributes} summarizes their key attributes. Moirai is an encoder-only Transformer trained on 27B observations using mixture log-likelihood loss. It supports multivariate input and models cross-input dependencies via attention, using multi-resolution tokenization (at multi-time scales) over fixed-length patches of a sequence. In contrast, TimesFM is a decoder-only Transformer trained on 100B+ time points (including LOTSA dataset) across energy, weather, and traffic domains. It is univariate but incorporates covariates (auxiliary inputs) and models TS patches via residual blocks, creating tokens to forecast.  Similarly, Chronos \cite{ansari2024chronos} is a univariate, pretrained TSFM that tokenizes continuous values into discrete bins and uses a transformer to predict future sequences autoregressively. Its Chronos-X\cite{chronosX} extension incorporates exogenous covariates through cross-attention mechanisms allowing the model to leverage external variables while maintaining strong zero-shot forecasting capabilities.
\noindent\textbf{\textit{C) Spatio-Temporal Graph Neural Networks (STGNNs): }}STGNN models forecast time series by capturing temporal patterns and spatial dependencies using a graph structure, where an adjacency matrix encodes node interactions. STGNNs operate in two learning stages: (i) Temporal learning, where each node processes its historical data using sequential neural network models like Gated Recurrent Units (GRU) or Convolutional Neural Network (CNN) to capture local temporal patterns and (ii) Spatial aggregation, where nodes exchange information with neighbors via Graph Convolutions Network (GCN), with edge weights reflecting inter-node relation. Each node updates its state by combining local and neighbor features. We use 2 STGNNs: \textbf{(a) GRUGCN \cite{GRUGCN}-} first processes each node's input through GRU units to capture temporal patterns, then applies graph convolutions to aggregate spatial information from neighbor nodes. \textbf{(b) TGCN \cite{TGCN}-} applies a two-layer GCN to the raw input features at each time step to model spatial dependencies, and then uses a GRU to capture temporal patterns. Unlike GRUGCN, the GCN in T-GCN operates on input features rather than GRU outputs. This allows earlier incorporation of neighborhood information. The choice between them depends on whether spatial dependencies should inform temporal processing (TGCN) or follow it (GRUGCN).
\vspace{-0.7cm}
\begin{table}[H]
\small
\renewcommand{\arraystretch}{0.95}
\setlength{\tabcolsep}{3pt}
\centering
\caption{\small Time Series Foundation Models attributes}
\vspace{-0.4cm}
\label{tab:tsfm-base-attributes}
\resizebox{\columnwidth}{!}{%
\begin{tabular}{@{}llll@{}}
\toprule
\textbf{Attribute} & \textbf{Chronos (Amazon)} & \textbf{TimesFM (Google)} & \textbf{Moirai-Small (Salesforce)} \\
\midrule
Zero-shot Capability & \checkmark & \checkmark & \checkmark \\
Multivariate Forecasting & \ding{55} (univariate) & \ding{55} (univariate, adapted) & \checkmark (any-variate) \\
Covariate Support & \checkmark (via ChronosX) & \checkmark & \checkmark \\
Forecasting Type & Probabilistic (quantile) & Point-wise & Probabilistic \\
Irregular TS Handling & \ding{55} & \ding{55} & \checkmark \\
Pretraining Corpus Size & 4B time-points & 100B time-points & 27B observations (LOTSA) \\
Primary Domains & General purpose & Finance, Climate & Finance, Climate, Transport \\
Architecture & Encoder–Decoder (T5) & Decoder-only Transformer & Encoder-only Transformer \\
Tokenization Strategy & Quantile-based binning & Patch-wise (fixed) & Patching + multi-resolution \\
Temporal Shift Robustness & Moderate & \ding{55} & \checkmark (via normalization) \\
Computational Cost & Quadratic (self-attn) & Quadratic (self-attn) & Quadratic (self-attn) \\
\bottomrule
\end{tabular}%
}

\end{table}
\vspace{-0.4cm}

\vspace{-0.4cm}
\section{Methodology}
\vspace{-0.1cm} We evaluate forecasting performance on meteorological sensor network data, where each model predicts future temperature readings given historical inputs. Our framework maintains fixed durations (8h context window $W$, 4h prediction horizon $P$) while varying: (1) sampling rates $f_s \in \{5,15,30,45,60\}$ minutes, and (2) network density with node counts $K  \in \{8,16,25\}$ representing progressively larger spatial coverage - 8 nodes form the most tightly clustered subgroup, 16 nodes cover intermediate proximity, and 25 nodes encompass the full network. We calculate the context length in samples as $C = 60 \times W/f_s = 480/f_s$, yielding a 96 context samples at the finest 5-minute resolution ($f_s=5$), decreasing to 8 samples at the coarsest 60-minute resolution ($f_s=60$). The prediction length follows similarly as $H = 60\times P/f_s = 240/f_s$, producing forecast sequences ranging from 48 steps (5-minute) down to 4 steps (60-minute) per sensor. This creates a continuous temporal resolution spectrum from 5-minute to 60-minute sampling. Models process spatial relationships differently: Moirai handles all nodes as multivariate inputs, while TimesFM and Chronos treat each sensor as an independent univariate series with forecasts aggregated via ensemble averaging. STGNNs construct graphs using absolute Pearson correlations ($|\rho_{ij}|$) to capture micro-climatic patterns beyond physical proximity. 
This design enables comprehensive evaluation across temporal granularities, spatial coverage densities across model types.\vspace{-0.26cm}
\noindent\paragraph{Experimental design} We used the Internet-of-Battlefield Things (IoBT) data set \cite{iobt}, containing meteorological data (temperature, humidity, rainfall, etc.) from 25 sensor towers covering a 10km x 10km area in New Mexico, sampled every 5 minutes over 9 days. The high sampling frequency provides substantial data volume for short-term forecasting evaluation, with each sensor contributing $9 \times 24 \times 12 = 2,592$ samples, totaling 64,800 time steps across all sensors. This volume is well-suited for short-term forecasting where models rely on recent temporal contexts (predicting 4H from 8H histories), and the effective sample size is further amplified through sliding window processing, generating thousands of training sequences for hour-ahead prediction. The dataset is split into 5 training, 2 validation, and 2 testing days, consistent with standard practices for short-term forecasting evaluation \cite{stgnn,timesFM}. Temperature forecasting performance is reported as MAE and RMSE averaged across nodes.  The sensor network layout and spatial node coverage for $K\in\{8,16,25\}$ nodes are provided in Appendix A.
\vspace{-0.25cm}
\noindent\paragraph{Enhancing TimesFM and Chronos with spatial interdependency modeling} To capture cross-sensor dependencies, we extend Chronos X and TimesFM with a post-hoc ensemble similarity method. For each target sensor, we select its top-k most correlated neighbors (using Pearson similarity) and blend the target’s forecast with neighbors’ predictions using similarity-based weights, effectively treating neighboring forecasts as exogenous covariates. This adaptive blending exploits spatial relationships to improve robustness under heterogeneous sensor conditions while preserving both models’ zero-/few-shot behavior.
\vspace{-0.35cm}
\subsection{Model Building and Implementation}
\vspace{-0.1cm}
\subsubsection{Baseline models and TSFMs.} We evaluate 3 classical models as baselines— VAR, GRU, and Transformer— using the window and horizon configurations aforementioned. VAR operates directly on raw multivariate data without training, whereas GRU and Transformer employ trained 2-layer encoders (hidden size=64) on fixed windows to match zero-shot evaluation conditions. For TSFMs, we use Moirai, TimesFM, and Chronos, all evaluated in zero-shot inference mode without any fine-tuning. Moirai processes all nodes jointly as a $K \times W$ matrix, using self-attention to capture cross-sensor dependencies and output probabilistic forecasts. In contrast, TimesFM and Chronos employ an ensemble similarity approach: each node is forecasted independently using its historical context, then predictions are blended via correlation-weighted averaging of the top-$k$ most similar neighbor forecasts (with $k=3$ and $\alpha=0.6$ weight for the target node).
\vspace{-0.2cm}
\subsubsection{STGNNs.} Graph construction is critical in STGNNs, as it defines the structure ($\mathcal{G} = (\mathcal{V}, \mathcal{E})$, where $\mathcal{V}$ and $\mathcal{E}$ represent the sets of sensor nodes and weighted connections between them, respectively). The goal is to model relationships between nodes (sensors) based on their time series data, accounting for spatial correlations due to shared micro-climate factors (e.g., altitude, local events) and temporal variations due to multi-frequency sampling (e.g., 5-min vs. 60-min intervals). In order to achieve a head-on-head comparison with other forecasting models, we aggregate raw time series data sequences from sensors onto a common timeline, build $\mathcal G$ and then run a standard \emph{window\,$\to$\,horizon}
forecasting pipeline. We construct the graphs using a Pearson correlation for building an adjacency matrix under varying redundancy levels ($p\%$). Here, redundancy is the degree of shared information between sensors due to spatial correlations. Controlling redundancy balances noise reduction (via correlated signals) and information diversity (via unique signals), optimizing the graph’s ability to capture meaningful spatial relationships without over-fitting or introducing overly sparse connections missing important details. To construct the adjacency matrix, for $N$ sensors with time series $X_i = [X_{1,i},...,X_{T,i}]^\top$, we compute the absolute Pearson correlation $\rho_{ij} = |\text{corr}(X_i,X_j)|$ between each pair $(i,j)$. This produces a symmetric $N \times N$ matrix where each entry quantifies the strength of data correlation between sensor pairs. To optimize the graph’s predictive utility, we threshold these correlations by retaining only the top $p\%$ of connections. The threshold $\theta$ is adaptively determined based on the desired redundancy level $p$:
\vspace{-0.1cm}
\begin{equation}
\theta = \begin{cases}
\max \mathcal{U} & \text{if } p = 0\% \text{ (no edges)} \\
\text{percentile}(\mathcal{U},100-p) & \text{if } 0 < p < 100\% \\
\min \mathcal{U} & \text{if } p = 100\% \text{ (full connectivity)}
\vspace{-0.1cm}
\end{cases}
\end{equation}
where $\mathcal{U} = \{|\rho_{ij}| : i < j\}$ is the set of unique correlations and $p$ is the desired redundancy level (working example provided in Appendix). This approach preserves the most significant micro-climatic relationships while avoiding over-connection, as visualized by heatmaps for adjacency matrices in Fig. 1 for $p=20\%$ and $p=60\%$ redundancy levels, showing how increasing redundancy levels gives denser graph network connectivity.
\vspace{-0.2cm}
\begin{figure}[t]
  \centering
  \includegraphics[width=0.35\columnwidth]{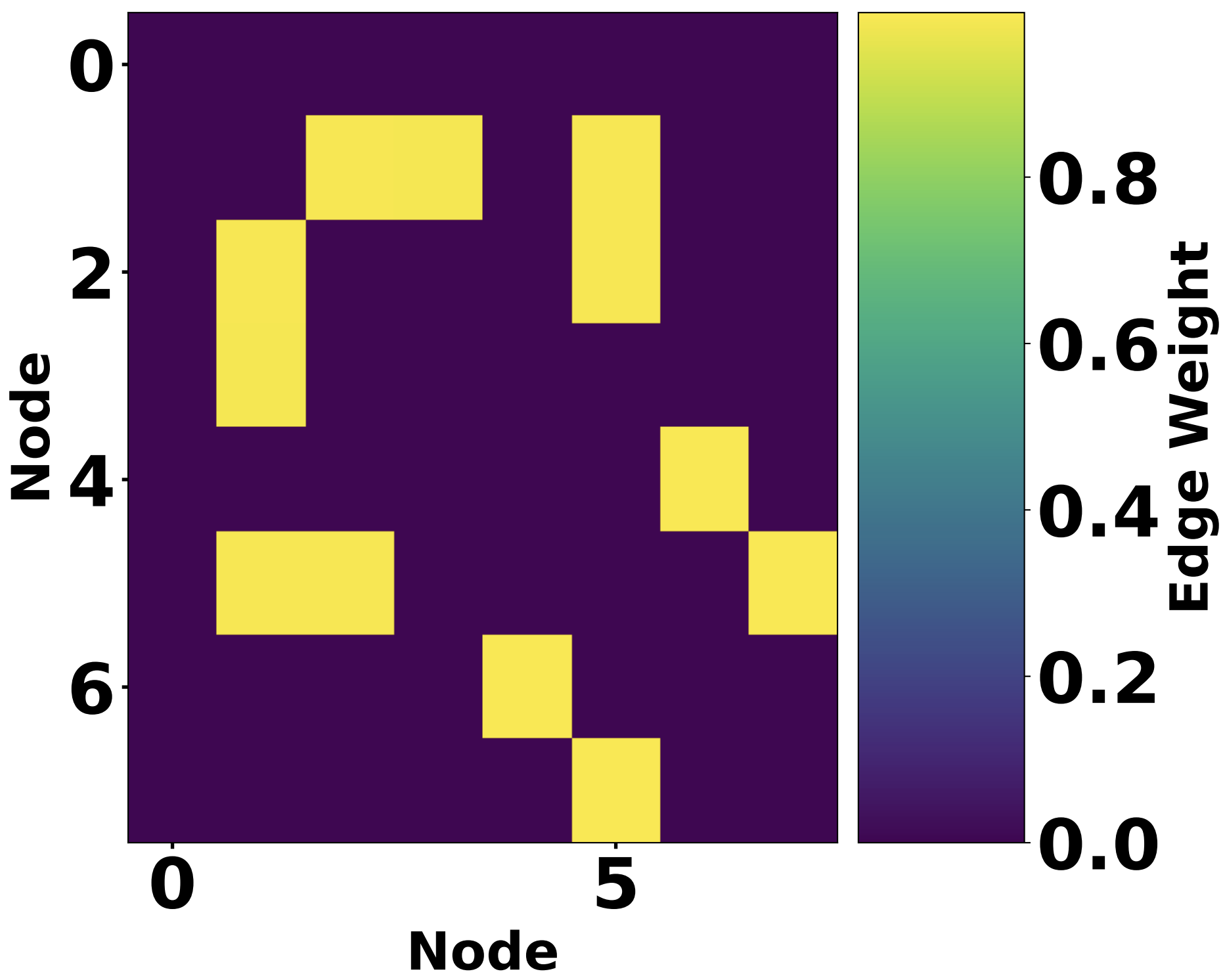}
  \includegraphics[width=0.35\columnwidth]{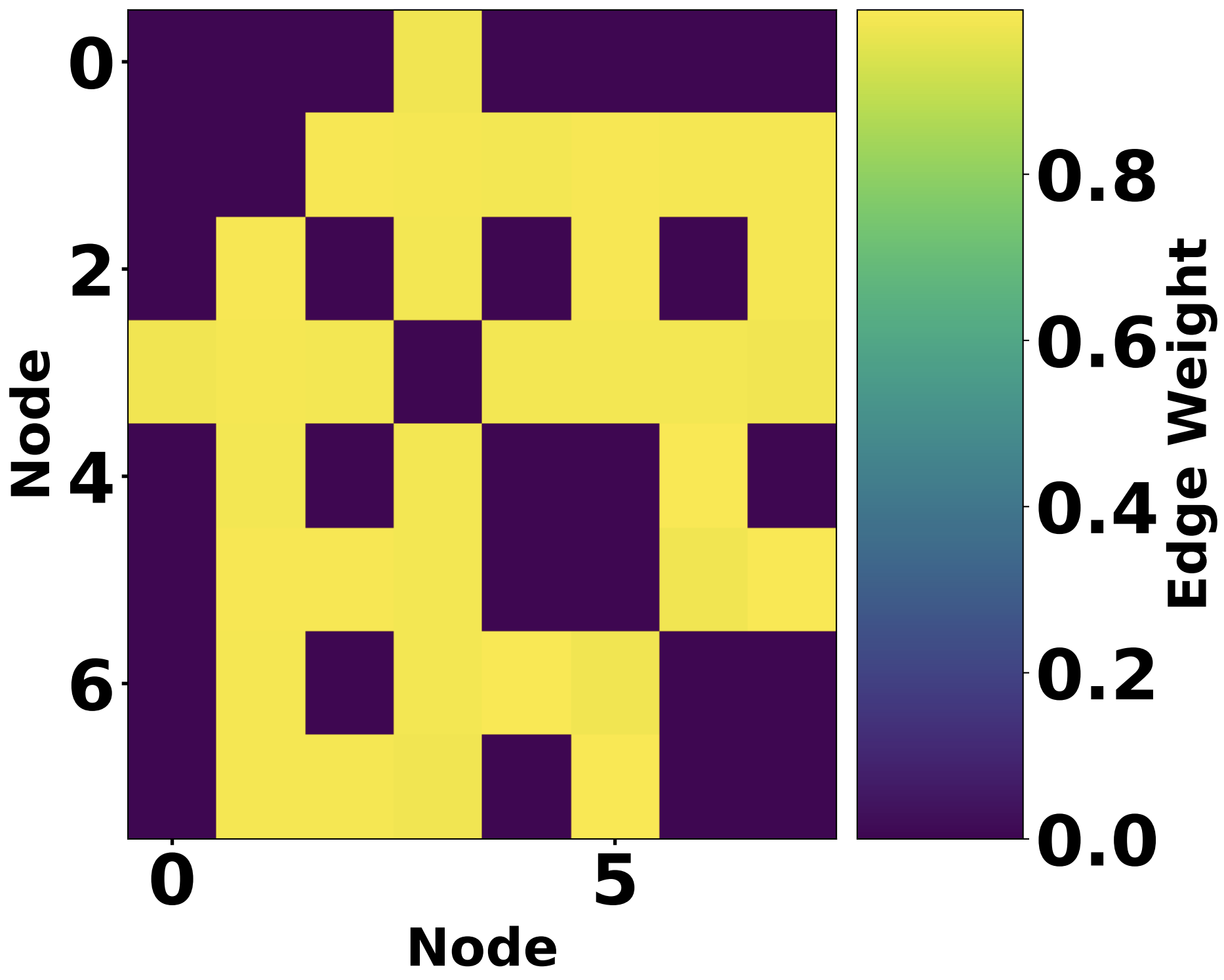}
  \caption{\small Adjacency matrix heatmaps for $p=20\%$ (left) and $p=60\%$ (right) in STGNNs.}
  \vspace{-0.2cm}
  \label{fig:metrics_visual}
\end{figure}

\subsubsection{Implementation details.} GRUGCN and TGCN are implemented using the Torch Spatiotemporal (TSL) framework with PyTorch Lightning. Both use one GRU and one Graph Convolutional Network (GCN) layer. GRUGCN processes input tensors of shape $(B, W, N, 1)$, where $B$ is batch size, $W$ is window length, and $N$ is the number of nodes (same as $K$). Models are trained with the Adam optimizer (learning rate $1 \times 10^{-3}$) and a batch size of 64.
\vspace{-0.3cm}

\section{Experimental Results}
\vspace{-0.2cm}
\noindent\paragraph{A. Baseline models and TSFMs.}
\vspace{-0.1cm}
Forecast performance for baseline models and TSFMs appears in Tables 2 and 3, respectively.

\noindent\textbf{(1) Impact of sampling rate (temporal granularity).} VAR excels at slow sampling (15–60 min) because linear cross-series structure dominates, but adds little at 5 min where dense temporal signals already suffice. GRU worsens at mid-range (15–45 min) as it struggles to trade off temporal vs. spatial cues, yet improves at 60 min when spatial correlations become clearer. Transformer peaks at the extremes—5 min (rich temporal detail) and 60 min (strong global patterns)—but weakens in the mid-range where neither signal is dominant.
When compared to TSFMs, the advantages of pretrained architectures become clear. At high frequency (5 min), Moirai moderately outperforms both baselines and other TSFMs, while at low frequency (60 min) its strength is pronounced (e.g., MAE = 0.93 with 8 nodes vs. 2.98 for GRU and 2.53 for Transformer), showing the benefit of multi-scale temporal representations that exploit long-range dependencies when temporal data is sparse. TimesFM degrades at coarser rates (e.g., RMSE $\geq$ 25 at 60 min/8 nodes) because its autoregressive token generation accumulates errors without global context. In contrast, Chronos is consistently more robust—e.g., MAE = 3.810 at 15 min/16 nodes (65\% lower than TimesFM’s 11.034) and remains stable at coarser rates. Post-hoc ensemble similarity with covariates narrows this gap where it reduces TimesFM errors by 11–12\% at 15–30 min/16 nodes but yields only 1.4\% reduction for Chronos. This is attributed to TimesFM's decoder-only architecture that generates tokens autoregressively, which struggles to capture extended temporal dependencies leading to performance degradation at longer horizons (30-60 minutes). Therefore, neighbor-based smoothing corrects these predictions significantly. Conversely, Chronos’s predictions are already spatially coherent (given it's encoder-decoder tokenization). It is worth noting, while ensembling can stabilize TimesFM, the most sampling-robust performance comes from models that natively learn spatial dependencies (in their core transformer architecture) with Moirai. TSFM forecast graphs at the lowest (5 min) and highest (60 min) sampling rates are shown in Appendix A.

\noindent\textbf{(2) Effect of cluster size (spatial node coverage).}
Model rankings shift with the number of nodes. For small clusters (\(8\) nodes), lightweight baselines can still be competitive; for example, at \(15\)-min/\(8\) nodes a plain Transformer slightly outperforms GRU and VAR (MAE \(2.02\) vs.\ \(2.51\) and \(2.38\)), showing that attention can capture short-range structure even in small systems. As clusters grow to \(25\) nodes, architectures that better exploit cross-series interactions gain an edge, wherein the Transformer improves or holds steady (e.g., \(5\)-min MAE \(\sim2.15\rightarrow\sim2.07\) from \(8\rightarrow25\) nodes), while GRU and VAR either plateaus or worsen (GRU \(2.16\rightarrow2.46\); VAR \(2.48\rightarrow2.57\)). A notable exception is VAR’s strong performance at \(45\)-min/\(25\) nodes (MAE \(1.95\)), where shared linear structure across many series is efficiently exploited. When compared with TSFMs, \textit{Moirai} is the strongest overall and even improves with scale (e.g., \(5\)-min MAE \(2.04 \rightarrow 1.85\) from \(8 \rightarrow 25\) nodes). Its a multi-resolution architecture pretrained on diverse LOTSA corpa benefits model predictions with increased node count adding rich spatial context. Its masked encoder excels at coarser sampling (30–60 min), where temporal signals are sparse, by extracting shared spatial patterns across sensors. In contrast, \textit{TimesFM} struggles on longer horizons (30–60 min) and is largely insensitive to node count, because its decoder-only, channel-independent architecture processes series separately and fuses late, limiting learned spatial interactions and extended temporal dependencies. Chronos is largely insensitive to node count where it is often slightly better at 25 nodes (e.g., improvements at 30/45/60-min). Ensemble similarity adds only ~1–2\% further gain unlike that for TimesFM
\begin{table}[t]
  \centering
    \vspace{-0.1cm}
  \caption{\small Forecast performance (MAE / RMSE) for baseline models.}
  \vspace{-0.4cm}
  \label{tab:err_baselines}
  \footnotesize
\begin{tabularx}{\columnwidth}{ccXXX}
    \toprule
    \textbf{Sampling Rate ($f_s$)} & \textbf{Nodes ($K$) } & \textbf{GRU} & \textbf{Transf.} & \textbf{VAR} \\
     \midrule
     5  &  8  & 2.16/3.10 & 2.20/3.41 & 2.48/3.48 \\
     5  & 16  & 2.35/3.59 & 2.20/3.46 & 2.53/3.44 \\
     5  & 25  & 2.46/3.52 & 2.15/3.31 & 2.57/3.45 \\
    15  &  8  & 2.51/3.72 & 2.02/3.12 & 2.38/3.29 \\
    15  & 16  & 2.23/3.40 & 2.15/3.36 & 2.15/3.04 \\
    15  & 25  & 2.31/3.39 & 2.07/3.31 & 2.14/3.07 \\
    30  &  8  & 2.18/3.31 & 2.15/3.25 & 2.60/3.48 \\
    30  & 16  & 2.12/3.15 & 2.13/3.30 & 2.12/2.97 \\
    30  & 25  & 2.11/3.19 & 2.15/3.10 & 2.07/2.96 \\
    45  &  8  & 2.16/3.14 & 2.12/2.96 & 2.59/3.43 \\
    45  & 16  & 2.06/2.96 & 2.61/3.57 & 2.02/2.79 \\
    45  & 25  & 2.14/3.11 & 2.09/2.92 & 1.95/2.74 \\
    60  &  8  & 2.98/4.00 & 2.53/3.18 & 3.37/4.21 \\
    60  & 16  & 2.71/3.35 & 2.64/3.53 & 2.53/3.33 \\
    60  & 25  & 2.71/3.41 & 2.08/2.82 & 2.23/2.91 \\
  
\bottomrule
  \end{tabularx}
\par\medskip\footnotesize
Note: Values are in \degree C.
\end{table}

\vspace{-0.3cm}
\begin{table}[t]
  \centering
      \vspace{-0.3cm}
  \caption{\small Forecast performance (MAE/RMSE) for TSFMs.}
  \vspace{-0.4cm}
  \footnotesize
  \setlength{\tabcolsep}{1.5pt}
  \resizebox{\columnwidth}{!}{%
 \begin{tabular}{p{0.8cm}p{0.6cm}lllll}
    \toprule
    \textbf{$f_s$} & \textbf{$K$} &
    \textbf{Chronos} &
    \makecell{\textbf{Chronos}\\\textbf{(w/ Covariates)}} &
    \textbf{TimesFM} &
    \makecell{\textbf{TimesFM}\\\textbf{(w/ Covariates)}} &
    \textbf{Moirai} \\
    \midrule
     5 &  8 & 3.842 / 4.420 & 3.768 / 4.186 & 6.389 / 8.663 & 3.753 / 5.054 & 2.039 / 2.955 \\
     5 & 16 & 3.731 / 4.331 & 3.728 / 4.280 & 6.859 / 9.271 & 4.301 / 5.735 & 1.868 / 2.304 \\
     5 & 25 & 3.437 / 4.004 & 3.413 / 3.953 & 6.957 / 9.441 & 5.143 / 6.506 & 1.854 / 2.308 \\
    15 &  8 & 3.699 / 4.208 & 3.768 / 4.186 & 9.348 / 11.276 & 6.776 / 8.594 & 1.570 / 2.079 \\
    15 & 16 & 3.810 / 4.376 & 3.756 / 4.247 & 11.034 / 13.272 & 9.694 / 11.570 & 1.475 / 1.730 \\
    15 & 25 & 3.446 / 3.996 & 3.400 / 3.892 & 9.816 / 12.118 & 8.814 / 10.662 & 1.743 / 2.265 \\
    30 &  8 & 3.507 / 4.121 & 3.565 / 4.053 & 22.931 / 24.410 & 17.762 / 20.022 & 1.384 / 1.772 \\
    30 & 16 & 3.566 / 4.129 & 3.503 / 3.992 & 22.445 / 23.772 & 20.009 / 21.622 & 1.386 / 1.765 \\
    30 & 25 & 3.322 / 3.865 & 3.275 / 3.770 & 21.038 / 22.961 & 19.134 / 20.977 & 1.393 / 1.725 \\
    45 &  8 & 3.159 / 3.781 & 3.046 / 3.614 & 8.224 / 10.207 & 5.991 / 7.066 & 0.861 / 1.130 \\
    45 & 16 & 3.173 / 3.762 & 3.105 / 3.641 & 7.788 / 9.473 & 6.669 / 7.603 & 1.221 / 1.424 \\
    45 & 25 & 2.975 / 3.581 & 2.940 / 3.501 & 7.301 / 8.998 & 6.455 / 7.517 & 0.957 / 1.252 \\
    60 &  8 & 4.042 / 4.505 & 4.145 / 4.478 & 22.382 / 25.823 & 22.051 / 25.513 & 0.932 / 1.196 \\
    60 & 16 & 4.154 / 4.581 & 4.273 / 4.611 & 21.468 / 25.045 & 21.111 / 24.592 & 0.862 / 1.263 \\
    60 & 25 & 3.922 / 4.405 & 3.987 / 4.410 & 21.587 / 25.146 & 21.354 / 24.802 & 0.881 / 1.168 \\
    \bottomrule
  \end{tabular}%
  }
  \par\medskip\footnotesize
Note: Values are in \degree C.

\end{table}

\begin{table*}[t!]
\centering
\small
\renewcommand{\arraystretch}{0.8}
\caption{Performance evaluation for STGNNs (GRUGCN and TGCN)}
\vspace{-0.4cm}
\label{tab:metrics_multirate}
\setlength{\tabcolsep}{2pt}
\begin{adjustbox}{max width=\textwidth}
\begin{tabular}{@{}r@{\hspace{2pt}}c@{\hspace{6pt}}cc cc cc cc cc | cc cc cc cc cc@{}}
\toprule
\multicolumn{2}{c}{\textbf{Hyperparameters}} &
\multicolumn{10}{c}{\textbf{GRUGCN}} &
\multicolumn{10}{c}{\textbf{TGCN}} \\
\cmidrule(lr){1-2} \cmidrule(lr){3-12} \cmidrule(l){13-22}
\textbf{Red.($p$)} & \textbf{Nodes($K$)} &
\multicolumn{2}{c}{\textbf{5 min}} &
\multicolumn{2}{c}{\textbf{15 min}} &
\multicolumn{2}{c}{\textbf{30 min}} &
\multicolumn{2}{c}{\textbf{45 min}} &
\multicolumn{2}{c}{\textbf{60 min}} &
\multicolumn{2}{c}{\textbf{5 min}} &
\multicolumn{2}{c}{\textbf{15 min}} &
\multicolumn{2}{c}{\textbf{30 min}} &
\multicolumn{2}{c}{\textbf{45 min}} &
\multicolumn{2}{c}{\textbf{60 min}} \\
\cmidrule(lr){3-4} \cmidrule(lr){5-6} \cmidrule(lr){7-8} \cmidrule(lr){9-10} \cmidrule(lr){11-12}
\cmidrule(lr){13-14} \cmidrule(lr){15-16} \cmidrule(lr){17-18} \cmidrule(lr){19-20} \cmidrule(l){21-22}
 &  & MAE & RMSE & MAE & RMSE & MAE & RMSE & MAE & RMSE & MAE & RMSE
 & MAE & RMSE & MAE & RMSE & MAE & RMSE & MAE & RMSE & MAE & RMSE \\
\midrule
\multicolumn{22}{l}{\textbf{0\%}}\\
\phantom{0\%} & 8   & 2.686 & 3.748 & 2.130 & 3.015 & 2.159 & 3.019 & 2.469 & 3.279 & 2.673 & 3.482
               & 2.568 & 3.700 & 2.289 & 3.505 & 2.008 & 2.975 & 2.013 & 2.848 & 3.246 & 4.156 \\
\rowcolor{blue!10}
\phantom{0\%} & 16  & 2.254 & 3.424 & 2.128 & 2.916 & 2.278 & 3.109 & 2.204 & 3.000 & 2.749 & 3.676
               & 2.684 & 3.700 & 1.829 & 2.638 & 2.171 & 2.992 & 2.164 & 2.899 & 3.033 & 3.881 \\
\phantom{0\%} & 25  & 2.343 & 3.461 & 2.309 & 3.196 & 2.283 & 3.020 & 2.306 & 3.075 & 2.886 & 3.590
               & 2.646 & 3.676 & 2.114 & 3.281 & 2.236 & 3.004 & 2.274 & 2.898 & 3.168 & 4.048 \\
\midrule
\multicolumn{22}{l}{\textbf{20\%}}\\
\phantom{20\%} & 8  & 2.328 & 3.442 & 2.209 & 3.069 & 2.296 & 3.075 & 2.262 & 2.987 & 2.668 & 3.306
                & 2.733 & 3.753 & 2.415 & 3.697 & 2.023 & 3.017 & 2.004 & 2.793 & 2.890 & 3.773 \\
\rowcolor{blue!10}
\phantom{20\%} & 16 & 2.290 & 3.504 & 2.333 & 3.250 & 2.144 & 2.899 & 2.253 & 2.987 & 2.519 & 3.227
                & 2.388 & 3.601 & 1.994 & 2.919 & 2.113 & 2.951 & 2.056 & 2.768 & 2.351 & 3.030 \\
\rowcolor{blue!10}
\phantom{20\%} & 25 & 2.658 & 3.600 & 2.218 & 3.105 & 2.253 & 3.086 & 2.308 & 3.117 & 2.528 & 3.263
                & 2.297 & 3.259 & 2.038 & 3.104 & 2.201 & 3.117 & 1.976 & 2.704 & 2.784 & 3.676 \\
\midrule
\multicolumn{22}{l}{\textbf{60\%}}\\
\rowcolor{blue!10}
\phantom{60\%} & 8  & 2.614 & 3.653 & 2.366 & 3.244 & 2.088 & 2.889 & 2.286 & 3.132 & 2.803 & 3.578
                & 2.801 & 3.823 & 2.191 & 3.279 & 2.125 & 2.904 & 2.286 & 3.035 & 3.185 & 4.085 \\
\rowcolor{blue!10}
\phantom{60\%} & 16 & 2.579 & 3.585 & 2.089 & 2.993 & 2.064 & 2.802 & 2.168 & 2.934 & 2.582 & 3.279
                & 2.701 & 3.693 & 2.082 & 3.403 & 1.956 & 2.792 & 2.136 & 2.867 & 2.969 & 3.863 \\
\phantom{60\%} & 25 & 2.557 & 3.461 & 2.014 & 2.725 & 2.144 & 2.873 & 2.212 & 2.896 & 2.479 & 3.165
                & 2.665 & 3.623 & 2.037 & 3.008 & 2.017 & 2.722 & 2.209 & 2.842 & 3.022 & 3.830 \\
\midrule
\multicolumn{22}{l}{\textbf{100\%}}\\
\phantom{100\%} & 8  & 2.758 & 3.770 & 2.244 & 3.117 & 2.404 & 3.272 & 2.046 & 2.752 & 2.585 & 3.211
                 & 2.679 & 3.727 & 2.255 & 3.601 & 2.071 & 3.027 & 2.180 & 2.902 & 2.952 & 3.818 \\
\phantom{100\%} & 16 & 2.601 & 3.558 & 2.082 & 2.948 & 2.193 & 2.993 & 2.175 & 2.961 & 2.441 & 3.186
                 & 2.687 & 3.672 & 1.970 & 2.871 & 2.335 & 3.451 & 2.233 & 2.992 & 2.470 & 3.264 \\
\phantom{100\%} & 25 & 2.684 & 3.739 & 2.111 & 3.013 & 2.192 & 2.999 & 2.019 & 2.707 & 2.373 & 3.018
                 & 2.466 & 3.758 & 1.844 & 2.623 & 2.080 & 2.867 & 2.165 & 2.872 & 3.312 & 4.157 \\
\bottomrule
\end{tabular}
\end{adjustbox}
\end{table*}

\noindent\paragraph{B. Performance analysis of STGNNs.}Table 4 reports STGNN performance along two dimensions: (i) the effect of graph redundancy among inter-dependent nodes, and (ii) the interplay between sampling rate and node coverage (number of nodes). Both GRUGCN and T-GCN show non-linear sensitivity to graph redundancy, with optimal performance at 20–60\% connectivity. Both GRUGCN and T-GCN show a non-linear response to graph redundancy because of a classic bias–variance / oversmoothing trade-off in message passing. With too few edges, the model underuses spatial signal (high variance); with too many edges (100\% redundancy), repeated aggregation amplifies noise and oversmooths node features, making neighbors indistinguishable and introducing spurious correlations—hence the large MAE increases (GRUGCN +32\%, T-GCN +68\%). The sweet spot (with 20–60\% connectivity) retains informative neighbors while limiting noise propagation, which is why GRUGCN hits MAE 2.088 at 30min/8 nodes (60\%) and T-GCN peaks at MAE 1.976 at 45min/25 nodes (20\%). A 16-node graph is consistently robust where it balances spatial coverage with manageable redundancy so the receptive field is rich but not saturated. GRUGCN, with its temporal-first design, performs best at mid-range sampling rates (15–45 min), where there is enough temporal resolution for the GRU to extract sequential patterns before spatial aggregation; at coarse rates (60 min), the weaker temporal signal reduces its effectiveness and makes it more sensitive to graph redundancy. In contrast, T-GCN’s spatial-first architecture excels at the extremes: at 5 min, early spatial aggregation smooths noisy fine-grained fluctuations, while at 60 min it compensates for sparse temporal context by exploiting inter-node correlations. Redundancy further interacts with graph size—helping smaller graphs (8 nodes) by enriching limited spatial input, but leading to oversmoothing and diminishing returns in dense graphs (25 nodes). Overall, performance is maximized when the model’s processing order (temporal-first vs. spatial-first) and graph connectivity are aligned with the temporal granularity of the data. 

When comparing across model classes, Moirai delivers the best sampling-robust performance, even in zero-shot settings, due to its any-variate attention mechanism that helps modeling multivariate dependencies without model fine-training or extra context via covariates. In terms of univariate TSFMs tested under zero-shot setting, Chronos is the most reliable baseline. Post-hoc correlation-weighted ensembling with neighboring sensors can move TimesFM closer to Chronos, but gains are modest and architecture-dependent. STGNNs can surpass univariate TSFMs when redundancy is tuned and GNN design is aligned with sampling rates and node coverage. Moreoever, Chronos-X and TimesFM with covariates show only marginal improvements over their base versions, underscoring that simple covariate injection is less effective than Moirai.

\vspace{-0.5cm}
\section{Discussion}
\vspace{-0.15cm}
\noindent{\textit{A. No-one-size fits all- model performance is conditional.}} Each
model class excels under specific conditions. STGNNs (GRUGCN, T-GCN perform well at mid-to-high sampling rates and moderate redundancy. They leverage spatial graphs effectively, especially
with 16 nodes where other models do not perform well (is this right).
However, at low sampling or extreme redundancy, they can overfit
or oversmooth. VAR model improves with more sensor nodes by
capturing linear correlations but cannot model nonlinear patterns.
TSFMs (e.g., Moirai, TimesFM) excel on dense, high-frequency, uniform data but struggle with low spatial diversity; STGNNs are stronger in sparse or highly correlated deployments.

\noindent{\textit{B. Redundancy and graph structure must be tuned, not maximized.}} In case of STGNN, we observe that moderate spatial redundancy (40-60\%) optimizes STGNN performance, while extreme sparsity (0-20\%) or density (80-100\%) degrades accuracy through underfitting or oversmoothing. This effect is most pronounced in small networks (8 nodes), where redundancy compensates for limited spatial coverage, but becomes counterproductive in large networks
(25 nodes) where excessive connections dilute local patterns. The architectural differences between GRUGCN and T-GCN further reveal that optimal redundancy depends on model design - GRUGCN’s
temporal-first approach benefits from moderate redundancy at medium sampling rates, while T-GCN’s spatial-first architecture struggles with sparse temporal signals at low frequencies. These findings emphasize that effective graph construction requires balancing connectivity with specificity. Neither maximally sparse nor fully dense graphs consistently outperform; instead, there exists a sweet spot in redundancy, typically around 40–60\%, that must be carefully matched to both network size and model architecture to achieve optimal performance.

\noindent{\textit{C. TSFMs show promise but need contextual grounding.}} Moirai's superior forecasting performance even under zero-shot settings, stems from its native multivariate design. Its any-variate attention mechanism flattens all sensor data into a single sequence, processed simultaneously by a masked encoder. This allows the model to directly learn cross-sensor dependencies through self-attention, using rotary position embeddings to preserve the temporal order of patches (of time series sequence) within this flattened sequence. This integrated approach to spatiotemporal learning is fundamentally more powerful than the late-stage covariate fusion or post-hoc ensembling used by univariate models, which merely append spatial context rather than learning it. Going forward, univariate TSFMs should become context-aware either by accepting natural-language prompts that describe situational factors or by adding architectural support for true multivariate inputs that learn cross time-series relationship.
\vspace{-0.5cm}
\section{Conclusion}
\vspace{-0.16cm}
Multivariate TSFMs achieve the best spatio-temporal forecasts by natively modeling cross-sensor relationships, surpassing GNNs. While contextual covariates offer slight gains for univariate TSFMs, well-tuned STGNNs still lead, highlighting the need for architectural improvements towards multivariate, context-aware foundation models.
\vspace{-0.7cm}
\section*{Acknowledgment}
\vspace{-0.18cm}
This work was supported by The Boeing Company under the University Master Research Agreement \#2021-STU-PA-404.
\setlength{\bibsep}{1pt plus 0.3ex}
\bibliographystyle{ACM-Reference-Format}
\bibliography{samples/sample-base}
\newpage

\captionsetup[figure]{font=small,skip=3pt}
\captionsetup[sub]{font=small,skip=2pt}

\appendix
\section*{Appendix}

\section{Sensor layout}
Sensors in our IoT network record temperature, humidity, rainfall, and pressure
at fixed sampling rates, producing timestamped time series per sensor. Multiple
sensors yield multivariate, spatiotemporal data. Figure A.1 shows the geographical topology of the sensors used in this experiment, whereas Figure A.2 is important in motivating the key concept of redundancy where we see the overlap of temperature data as recorded by two colocated (belonging to the same cluster) sensors.

\begin{figure}[H]
  \centering
  \includegraphics[width=\columnwidth]{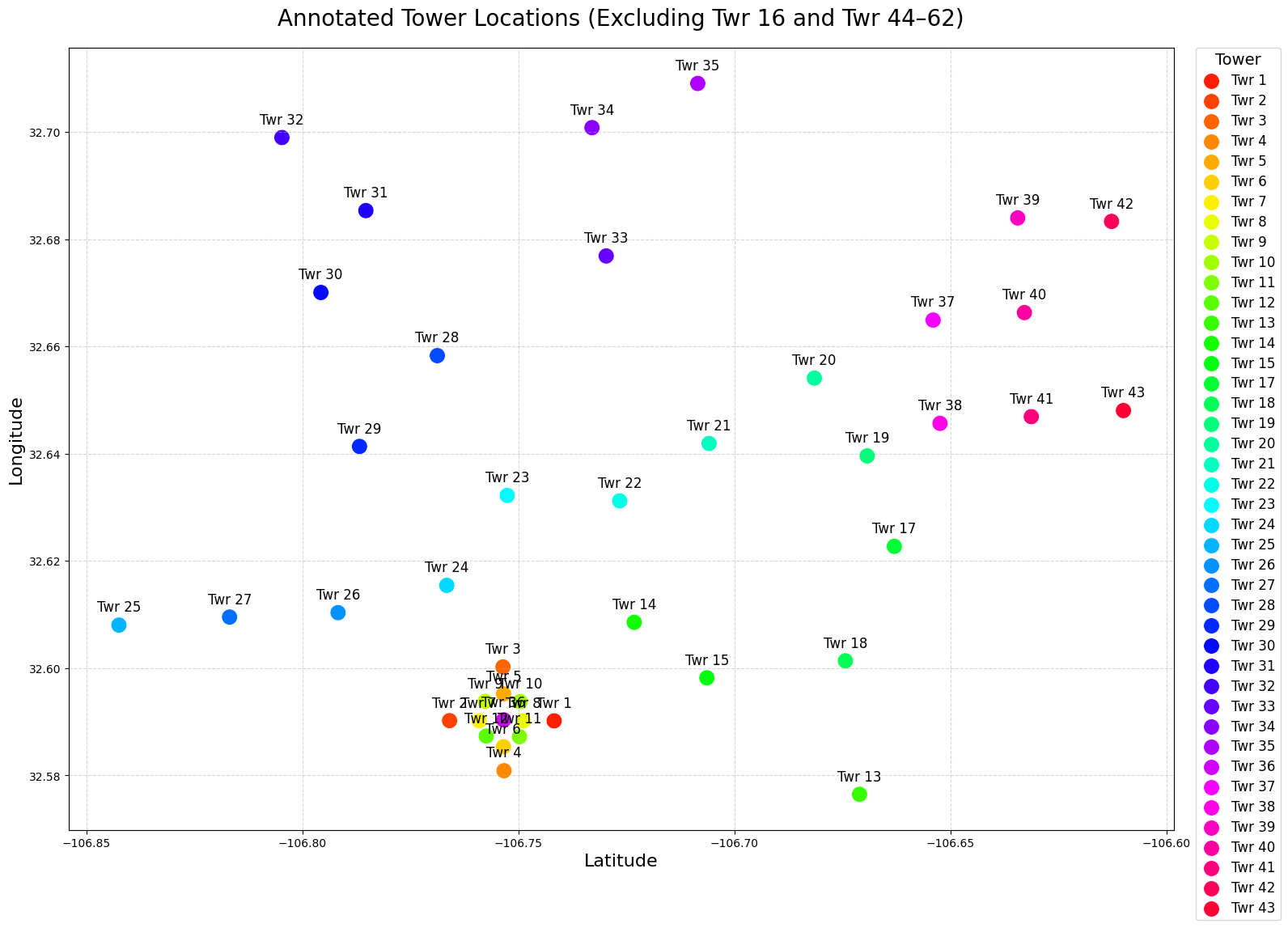}
  \caption{Annotated tower locations (lat--lon).}
  \label{fig:tower-map}
\end{figure}

\begin{figure}[H]
  \centering
  \includegraphics[width=\columnwidth]{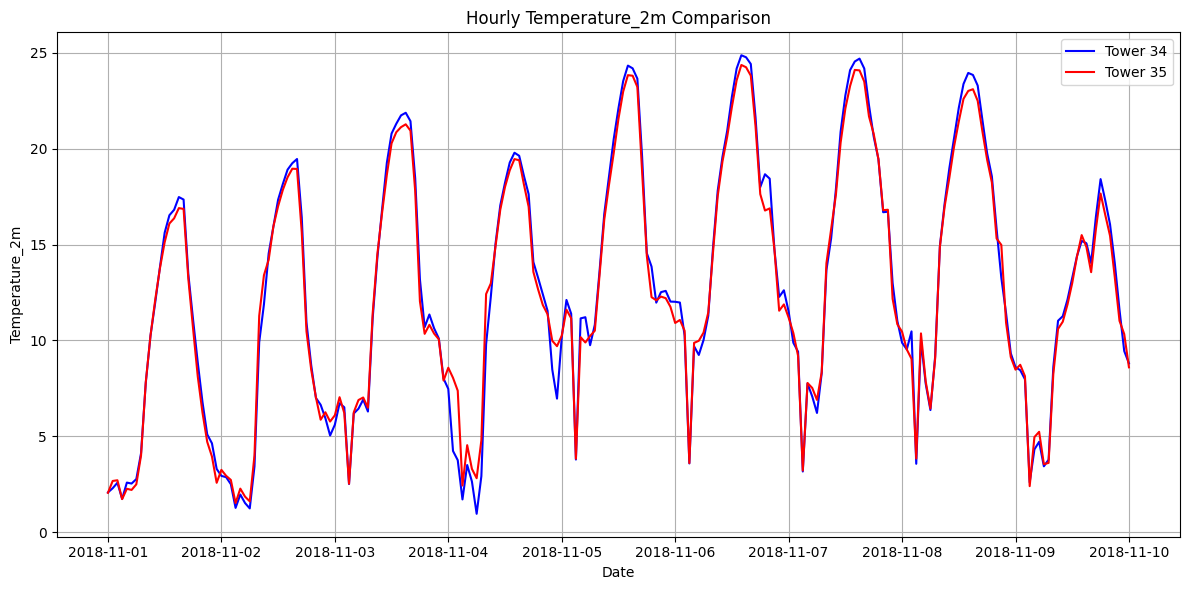}
  \caption{Hourly temperature overlap for colocated Towers 34 and 35.}
  \label{fig:temp-overlap}
\end{figure}

\section{Sensor selection for spatial node coverage}To select representative sensor subsets from the full 25-node network, we employ agglomerative clustering to ensure geographic representativeness. We partition all sensor locations into \(k=5\) clusters (treating each cluster as a Region of Interest- ROI), then select sensors for our subsets (\(K=8,16,25\)) through a systematic expansion process. Starting with the \(K=8\) subset, we choose sensors nearest to cluster centroids supplemented with geographically dispersed nodes. We then expand to \(K=16\) and \(K=25\) by iteratively merging neighboring clusters and incorporating their centroid sensors, maintaining representative spatial coverage while varying node density to evaluate model performance under different spatial configurations.
\vspace{-0.08em}
\setlength{\intextsep}{0pt}
\setlength{\textfloatsep}{0pt}

\begin{figure}[H]
  \centering
  \begin{subfigure}[t]{0.9\linewidth}
    \centering
    \includegraphics[width=\linewidth]{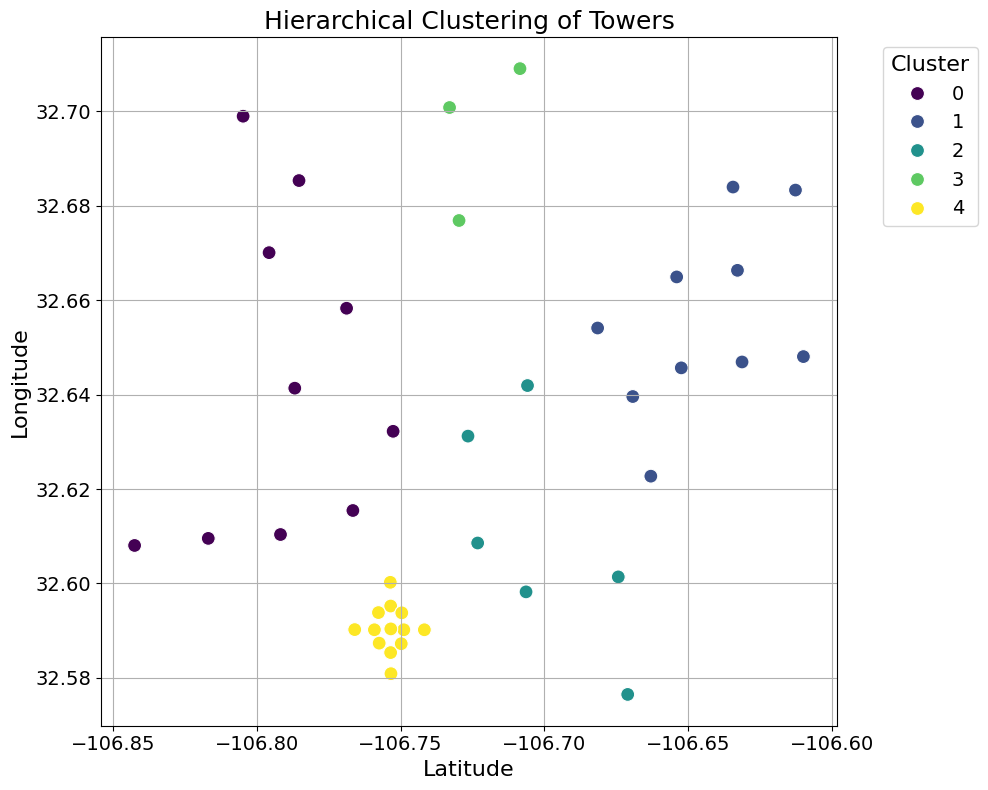}
    \caption{$k{=}5$ spatial clusters (colors).}
  \end{subfigure}

\vspace{1em}
  \begin{subfigure}[t]{0.9\linewidth}
    \centering
    \includegraphics[width=\linewidth]{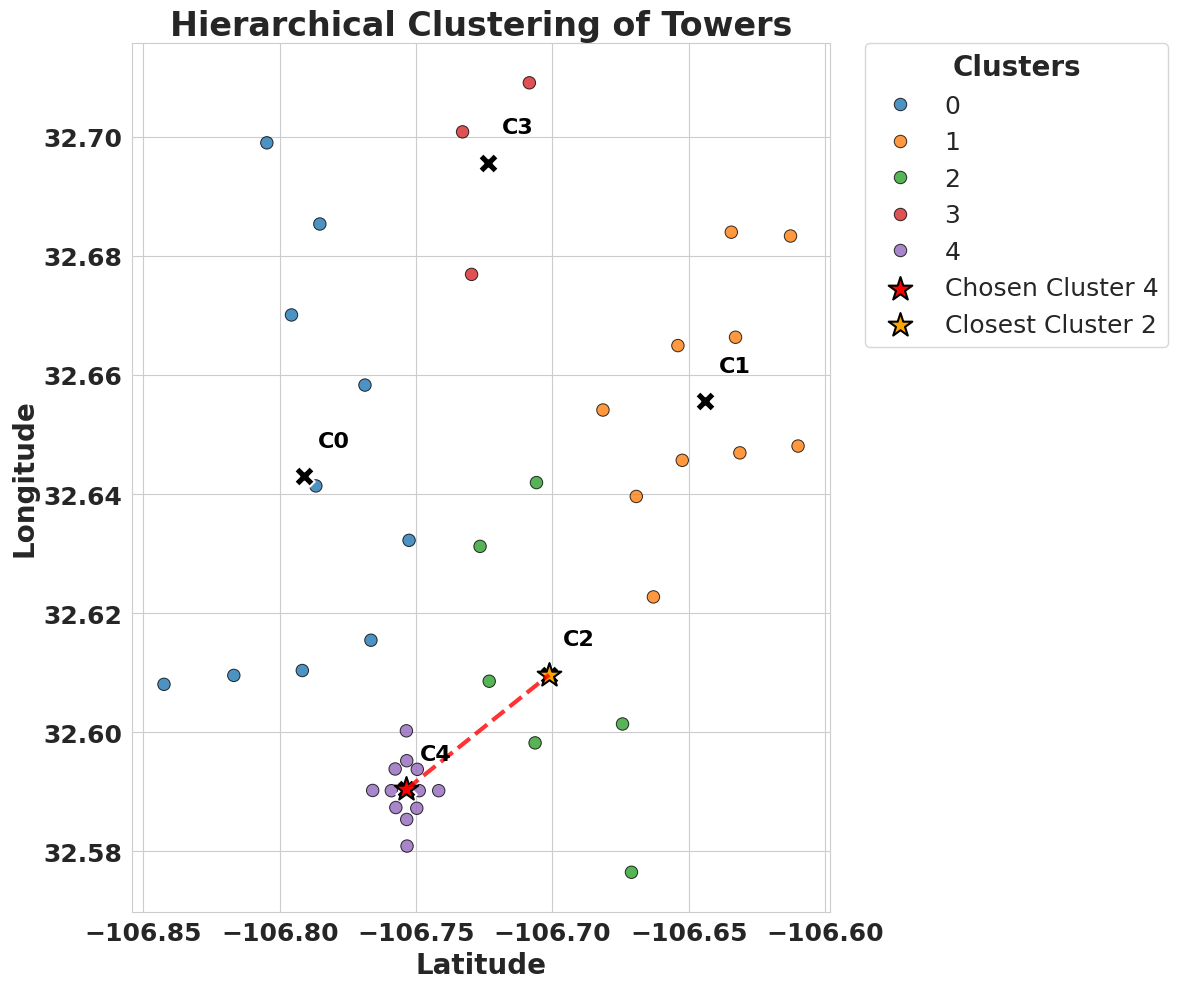}
    \caption{Cluster centroids; ROI's nearest neighbor.}
  \end{subfigure}
  \caption{(a,b) Spatial clustering and nearest-centroid selection.}
  \label{fig:spatial-clustering}
\end{figure}

\begin{figure}[H]
  \centering
  \begin{subfigure}[t]{0.9\linewidth}
    \centering
    \includegraphics[width=\linewidth]{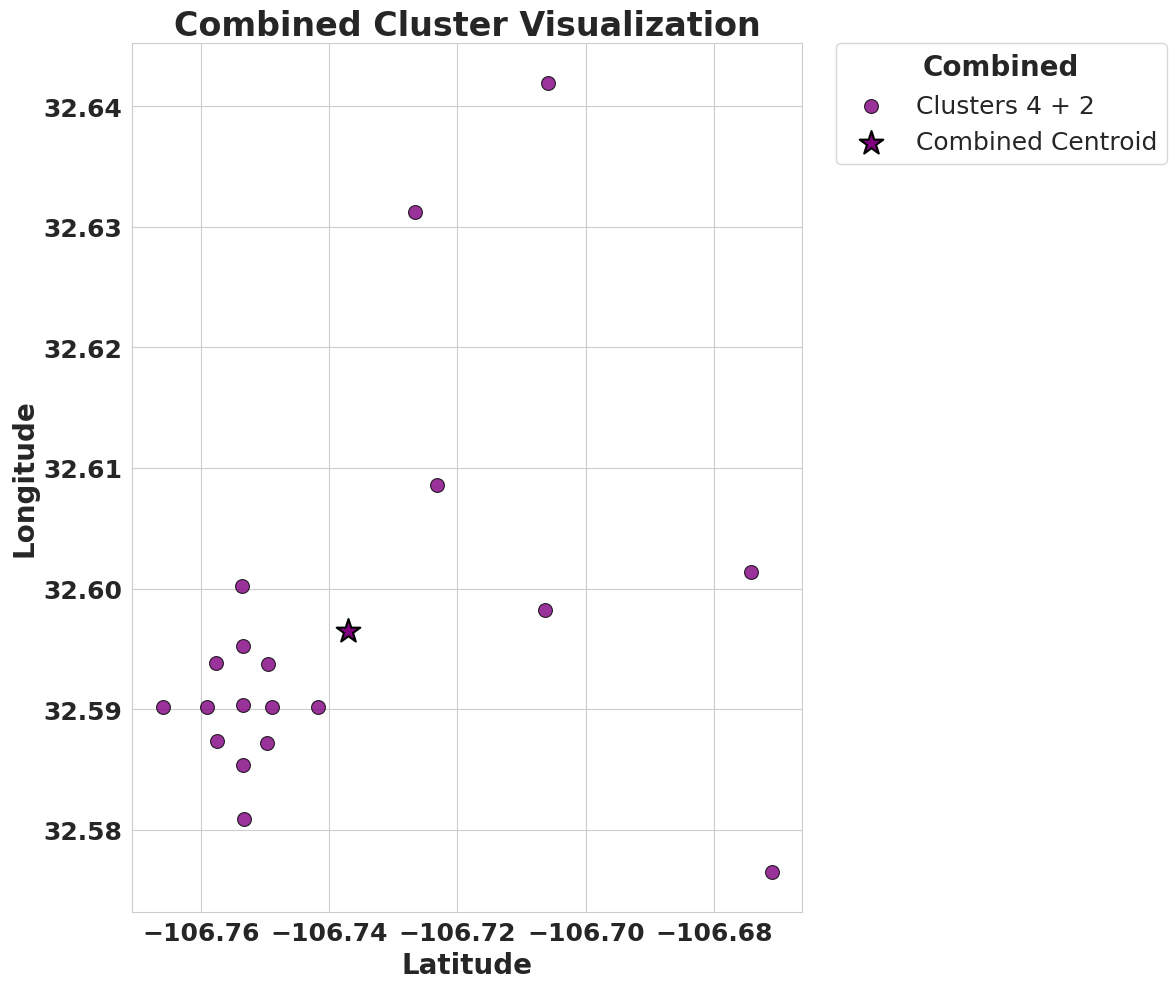}
    \caption{Merge ROI with nearest sensor tower (\ding{72} = merged centroid).}
  \end{subfigure}
\vspace{1em}
  \begin{subfigure}[t]{0.9\linewidth}
    \centering
    \includegraphics[width=\linewidth]{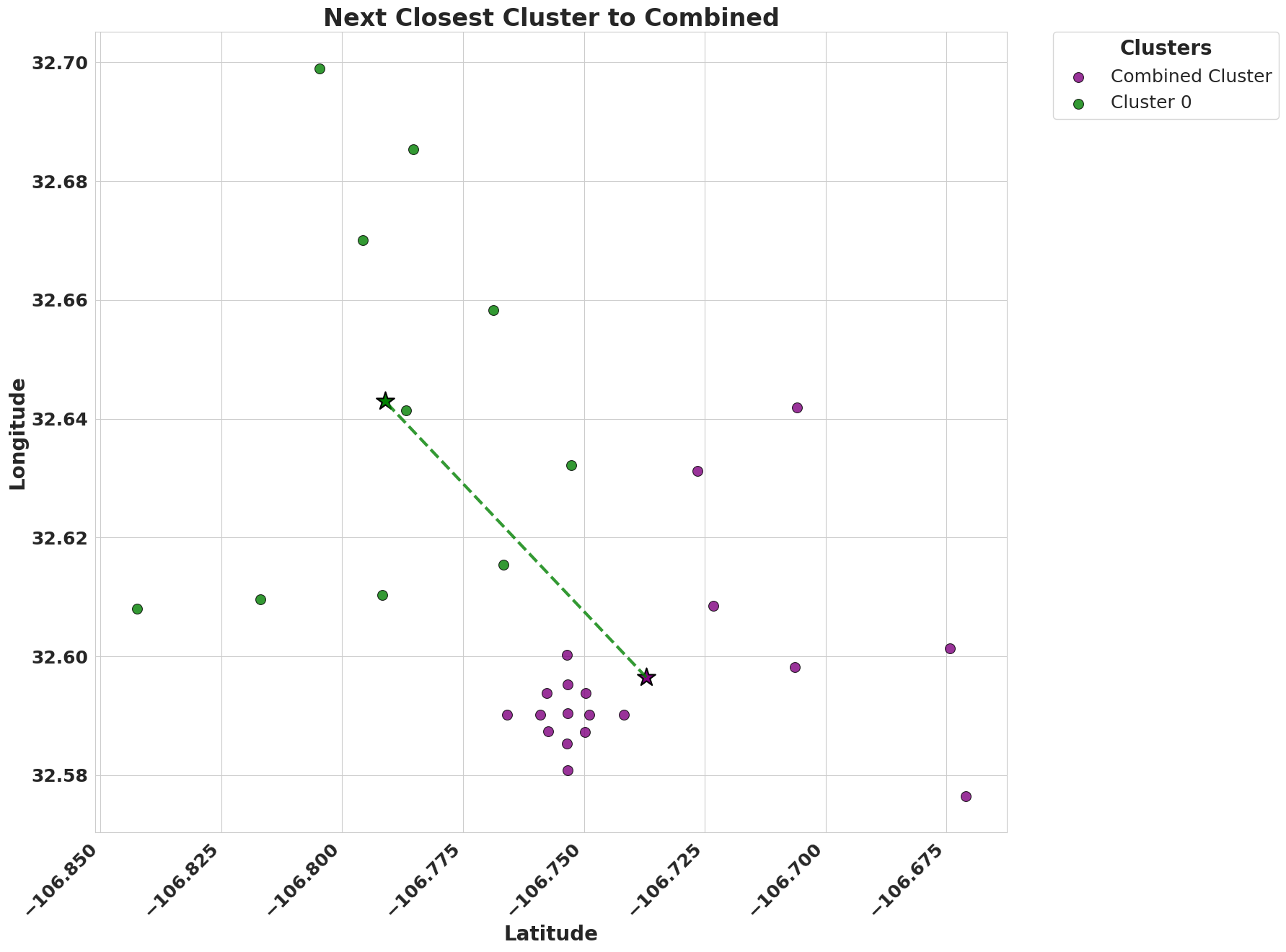}
    \caption{Next closest cluster to the merged centroid.}
  \end{subfigure}
  \caption{(c,d) Iterative merging of clusters by centroid distance.}
  \label{fig:merged-clustering}
\end{figure}

\section{A working example for modeling spatio-temporal dependencies in sensor network using GNNs} To demonstrate how a Graph Neural Network (GNN) models spatial dependencies among colocated sensors, we construct a small example using a \(5\times5\) Pearson correlation matrix.

\subsection*{Step 1: Pairwise correlation matrix}

We begin with a simple correlation matrix \( \rho_{ij} \) representing the Pearson correlation coefficients between five sensors measuring temperature:

\[
\rho_{ij} =
\begin{pmatrix}
1.00 & 0.85 & 0.78 & 0.62 & 0.55 \\
0.85 & 1.00 & 0.73 & 0.51 & 0.48 \\
0.78 & 0.73 & 1.00 & 0.66 & 0.59 \\
0.62 & 0.51 & 0.66 & 1.00 & 0.47 \\
0.55 & 0.48 & 0.59 & 0.47 & 1.00
\end{pmatrix}
\]

Here, the diagonal entries (\(\rho_{ii}=1\)) represent perfect self-correlation and are ignored when constructing the adjacency matrix. The matrix is symmetric, so we only need to consider the upper-triangular values (\(i < j\)) for thresholding. Extracting the upper triangle (excluding diagonal) yields:
\begin{align*}
\mathcal{R} &= \{0.85,\, 0.78,\, 0.62,\, 0.55,\, 0.73,\, 0.51,\, 0.48,\, 0.66,\, 0.59,\, 0.47\}, \\[1pt]
\mathcal{R}_{\mathrm{sorted}} &= \{0.47,\,0.48,\,0.51,\,0.55,\,0.59,\,0.62,\,0.66,\,0.73,\,0.78,\,0.85\}.
\end{align*}

\subsection*{Step 2: Thresholding by redundancy level}

To control redundancy, we retain only the top \(p\%\) of strongest correlations. This determines how densely connected the graph is.
\paragraph{For Redundancy \(\rho = 80\%\)}  
This implies the graph density  to retain the top \(80\%\) of correlations, i.e., the eight largest out of 10 values.  
The threshold becomes:
\[
\theta_{80} = \text{2nd-smallest value} = 0.48.
\]
Thus:
\[
A^{(80\%)}_{ij} =
\begin{cases}
\rho_{ij}, & \text{if } \rho_{ij} \ge 0.48,\\
0, & \text{otherwise}.
\end{cases}
\]

The resulting adjacency matrix is shown in Table C.1.

\begin{table}[H]
  \centering
  \caption{Adjacency at 80\% redundancy (\(\theta=0.48\)) based on Pearson correlation.}
  \setlength{\tabcolsep}{4pt}\scriptsize
  \begin{tabular}{c|ccccc}
  & 1 & 2 & 3 & 4 & 5\\
  \hline
  1 & 0 & \cellcolor{lightgray}0.85 & \cellcolor{lightgray}0.78 & \cellcolor{lightgray}0.62 & \cellcolor{lightgray}0.55\\
  2 & 0.85 & 0 & \cellcolor{lightgray}0.73 & \cellcolor{lightgray}0.51 & \cellcolor{lightgray}0.48\\
  3 & 0.78 & 0.73 & 0 & \cellcolor{lightgray}0.66 & \cellcolor{lightgray}0.59\\
  4 & 0.62 & 0.51 & 0.66 & 0 & \cellcolor{lightgray}0.47\\
  5 & 0.55 & 0.48 & 0.59 & 0.47 & 0
  \end{tabular}
\end{table}

Here, eight of ten unique pairs remain connected (\(8/10 = 80\%\)), producing a dense spatial graph. The adjacency matrix \(A\) defines how information flows between sensors in the GNN. The GNN uses this adjacency to perform graph convolution:
  \[
  \mathbf{H}^{(l+1)} = \sigma(\tilde{A}\mathbf{H}^{(l)}W^{(l)}),
  \]
  where \(\tilde{A}\) is the normalized adjacency derived from \(A\), ensuring that each node aggregates information proportionally to its correlations with neighbors. This process allows the GNN to learn spatial and temporal dependencies jointly: spatial through the correlation-based adjacency, and temporal through sequential updates across time windows. In our experiments, varying the threshold \(\theta\) effectively varies redundancy allowing us to quantify how connectivity strength among colocated sensors affects downstream forecasting accuracy.
\section{Time Series Foundation Models performance (5 minute and 60 minute sampling rates)}
An example visualization of our forecasting plots with context/history is shown as
solid gray, ground truth as solid blue, and model predictions as dashed lines
(green = MOIRAI, red = TimesFM, orange = Chronos). The vertical dotted line marks
the forecast start (split).
\begin{figure*}[t]
  \centering
  \begin{subfigure}[t]{0.48\textwidth}
    \centering
    \includegraphics[width=\linewidth]{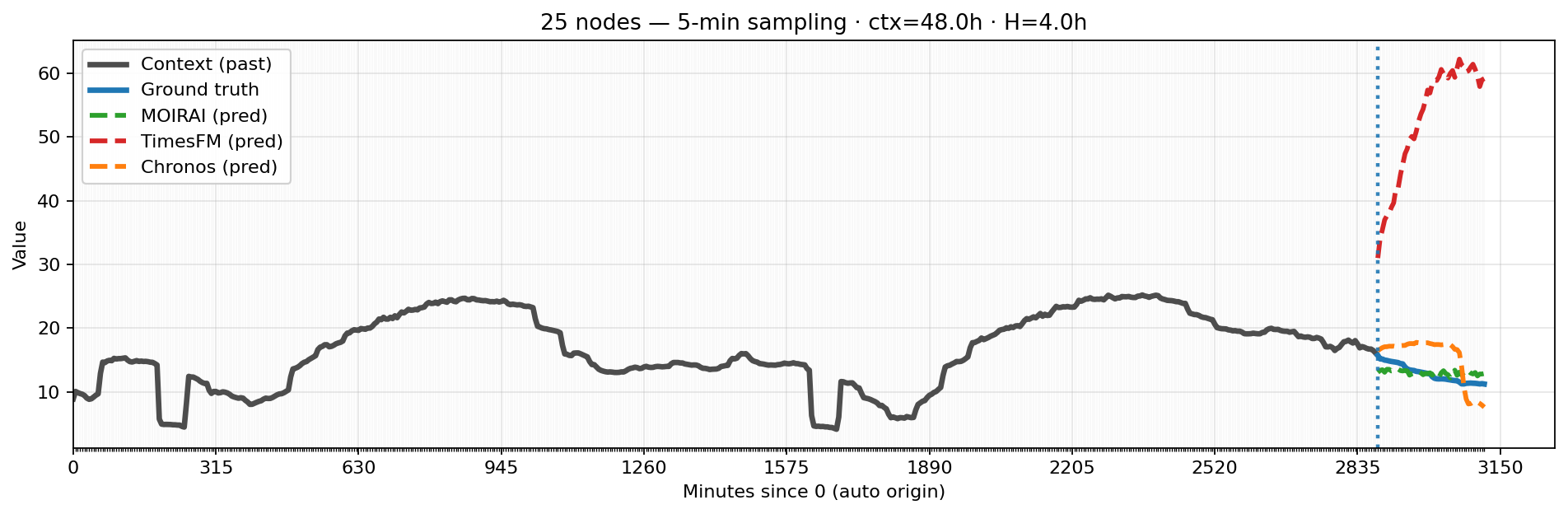}
    \caption{TSFM models forecasting at 5-minute sample rate.}
    \label{fig:tsfm-5}
  \end{subfigure}\hfill
  \begin{subfigure}[t]{0.48\textwidth}
    \centering
    \includegraphics[width=\linewidth]{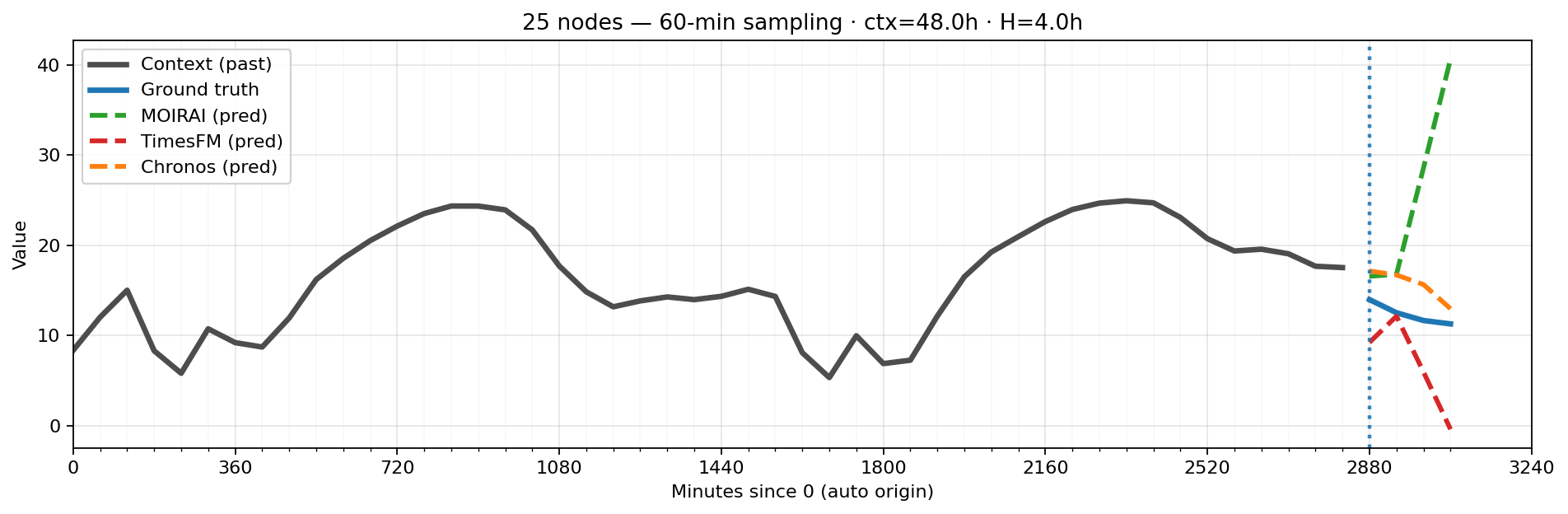}
    \caption{TSFM models forecasting at 60-minute sample rate.}
    \label{fig:tsfm-60}
  \end{subfigure}

  \caption{Comparing TSFM models' performance at different sampling intervals}
  \label{fig:tsfm-comparison}
\end{figure*}

\end{document}